\pdfoutput=1

\documentclass[11pt]{article}

\usepackage[preprint]{acl}

\usepackage{times}
\usepackage{latexsym}

\usepackage[T1]{fontenc}

\usepackage[utf8]{inputenc}

\usepackage{microtype}

\usepackage{inconsolata}

\usepackage{graphicx}
\graphicspath{ {figures/} }

\usepackage{multirow}
\usepackage{booktabs}
\usepackage{tabularx}
\title{Using LLMs for Automated Privacy Policy Analysis: Prompt Engineering, Fine-Tuning and Explainability}

\author{
  \textbf{Yuxin Chen\textsuperscript{1}}, 
  \textbf{Peng Tang\textsuperscript{1}}, 
  \textbf{Weidong Qiu\textsuperscript{1}},
  \textbf{Shujun Li\textsuperscript{2}}\\
  \textsuperscript{1} 
School of Cyber Science and Engineering, Shanghai Jiao Tong University, China\\
  \textsuperscript{2}Institute of Cyber Security for Society (iCSS), University of Kent, UK\\
    \texttt{qiuwd@sjtu.edu.cn, S.J.Li@kent.ac.uk}\\
    }

\begin{document}

\maketitle

\begin{abstract}
Privacy policies are widely used by digital services and often required for legal purposes. Many machine learning based classifiers have been developed to automate detection of different concepts in a given privacy policy, which can help facilitate other automated tasks such as producing a more reader-friendly summary and detecting legal compliance issues. Despite the successful applications of large language models (LLMs) to many NLP tasks in various domains, there is very little work studying the use of LLMs for automated privacy policy analysis, therefore, if and how LLMs can help automate privacy policy analysis remains under-explored. To fill this research gap, we conducted a comprehensive evaluation of LLM-based privacy policy concept classifiers, employing both prompt engineering and LoRA (low-rank adaptation) fine-tuning, on four state-of-the-art (SOTA) privacy policy corpora and taxonomies. Our experimental results demonstrated that combining prompt engineering and fine-tuning can make LLM-based classifiers outperform other SOTA methods, \emph{significantly} and \emph{consistently} across privacy policy corpora/taxonomies and concepts. Furthermore, we evaluated the explainability of the LLM-based classifiers using three metrics: completeness, logicality, and comprehensibility. For all three metrics, a score exceeding 91.1\% was observed in our evaluation, indicating that LLMs are not only useful to improve the classification performance, but also to enhance the explainability of detection results.
\end{abstract}

\section{Introduction}

In the digital age, the exponential growth of online services and applications has precipitated substantial concerns pertaining to user privacy protection. Some services or applications tend to excessively collect or utilize users' personal information, posing threats to privacy security. Privacy policies, serving as formal legal documents that delineate organizational data practices, constitute a critical mechanism for informing users about the collection, processing, storage and sharing of their personal data. The examination of privacy policies is of paramount importance for comprehending personal data processing mechanisms and evaluating organizational compliance with established privacy regulations such as the EU and the UK's GDPR (General Data Protection Regulation)~\cite{Voigt2017GDPR} and the USA's CCPA (California Consumer Privacy Act)~\cite{CCPA2018}. However, privacy policies are often complex, filled with technical terms, making comprehension challenging. Past research~\cite{ibdah2021why} has revealed that many users encountered difficulties in understanding the content of privacy policies. Therefore, analyzing privacy policies in a way that facilitates user understanding and comprehension holds significant practical value. Due to the increasing number of online services and applications and the iterative nature of privacy policies, manual analysis becomes unsustainable, making machine learning based automated privacy policy analysis a meaningful research direction.

Large language models (LLMs) have demonstrated the state-of-the-art performance in various natural language processing (NLP) benchmarks, showcasing a remarkable potential in practical applications such as text generation, dialogue processing, and knowledge question-answering~\cite{chang2024survey}. It is highly likely that LLMs will perform well in analyzing privacy policies written in natural language, as their capabilities can be effectively leveraged given the complex nature of these policies. Although many researchers have proposed machine learning based classifiers for automated privacy policy analysis, to the best of our knowledge, except the limited work by~\citet{goknil2024privacy} on exploring the use of prompt engineering LLMs for this purpose, the potential of LLMs remains largely unexplored.

In this paper, we report our comprehensive evaluation of utilizing both prompt engineering and LoRA (low-rank adaptation) fine-tuning to develop LLM-based privacy policy concept classifiers. We conducted experiments across four state-of-the-art (SOTA) privacy policy corpora/taxonomies and several mainstream LLMs, exploring the effects of different factors such as temperature and model size on the model performance. Explainability refers to the ability to explain or present the behavior of AI models in human-understandable terms~\cite{zhao2024explainability}. In addition to assessing the detection performance of LLM-based privacy policy concept classifiers, we also studied how to use LLMs to explain the detection results using customized prompts for different concepts. We evaluated the model's explainability across three metrics: completeness, logicality, and comprehensibility. Our key contributions are as follows:

\emph{1) We conducted a systematic evaluation on how to use LLMs to conduct automated privacy policy analysis, which, to the best of our knowledge, represents the first comprehensive study of this kind.} By leveraging both prompt engineering and LoRA fine-tuning, we managed to use LLMs to produce new privacy policy concept classifiers that can outperform other SOTA classifiers \emph{significantly} and \emph{consistently} across three mainstream open-source LLMs and four SOTA privacy policy corpora/taxonomies.

\emph{2) We systematically investigated the potential of using LLMs to explain detection results of LLM-based privacy policy concept classifiers.} Based on the above-mentioned three metrics, our human-based assessment results demonstrate that LLMs can generate meaningful explanations with high satisfaction (a score exceeding 91.1\% observed for all three metrics), although there are some shortcomings in logicality.

The remainder of this paper is structured as follows. Section~\ref{sec:related_work} presents related work. Section~\ref{sec:methodology} details our approach. Section~\ref{sec:experiments_results} outlines the experiment setup and results. Section~\ref{sec:explainability} explores the explainability of LLMs in privacy policy analysis. The last two sections conclude this paper and discuss limitations of the work, respectively.

\section{Related Work}
\label{sec:related_work}

\subsection{Privacy Policy Corpora}

Annotated privacy policy datasets are crucial for the training and evaluation of machine learning models. A common annotation involves segmenting privacy policies and classifying these segments based on taxonomies derived from legal standards or real-world privacy policies. As the first and the most widely used privacy policy dataset, OPP-115~\cite{opp-115} provides fine-grained annotations at the paragraph level. It encompasses 115 privacy policies from online services, with 3,792 paragraphs categorized into 12 privacy policy conceptual categories (which forms a mini-taxonomy). Each paragraph was independently annotated by three legal experts and assigned to one or multiple privacy policy concepts. Three more recent datasets were released in 2024. Among them, \citet{tang2024gdpr} introduced GoPPC-150, a dataset featuring paragraph-level annotations and a more comprehensive taxonomy tailored to GDPR requirements. Two other new datasets, CAPP-130~\cite{wen2024capp} and APPCP-100~\cite{zhang2024appcp}, focus on Chinese privacy policies, offering support for research on privacy policy analysis in a multilingual context.

\subsection{Automated Privacy Policy Analysis}

Automated privacy policy analysis encompasses various tasks such as concept classification~\cite{Srinath_2021, mousavi2020baseline, tang2024gdpr}, summary generation~\cite{wen2024capp}, question answering~\cite{harkous2018polisis}, and the annotation of key information like opt-out options~\cite{bannihatti2020finding}. Among these, concept classification in privacy policies has been more extensively studied. It involves segmenting privacy policies and labeling each segment based on a taxonomy covering relevant concepts. This approach can facilitate readers with a quick understanding of key conceptual points in different parts of the privacy policy. In addition, the coverage of concepts serves as an important criterion for assessing a privacy policy's legal compliance against a given data protection law.

Some researchers~\cite{torre2020ai, mousavi2020baseline, mustapha2020xlnet, Srinath_2021, tang2024gdpr} employed NLP approaches to automatically analyze the content of a given privacy policy and evaluated it on privacy policy corpora, establishing a stable baseline. Some others~\cite{xiang2023policychecker, cejes2024compai} adopted semantic role based approaches to do large-scale privacy policy completeness violation studies. However, there has been very little research on automated privacy policy analysis based on LLMs. The only past study we are aware of was done by~\citet{goknil2024privacy}, who looked at using prompt engineering LLMs for this purpose only.

\subsection{Large Language Models}

Large language models (LLMs), like OpenAI's GPT series~\cite{radford2018improving} and Meta's Llama series~\cite{touvron2023llama}, possess immense parameter sizes and learning capabilities. A notable capability of LLMs is their rich contextual learning ability~\cite{brown2020language}. Through carefully designed prompts, such as detailed task-specific instructions or a few illustrative examples, researchers can effectively guide models to generate targeted outputs. Many prompt engineering methods for LLMs have been developed in the past a few years~\cite{schulhoff2024promptreportsystematicsurvey}, e.g., \citet{Wei2022ChainOT} introduced the chain-of-thought (CoT) approach, which decomposes complex problems into intermediate reasoning steps, helping LLMs generate more logical and coherent responses. In addition to prompt engineering, which is more in the domain of zero- or few-shot training, fine-tuning is another effective way to improve LLMs' abilities of solving new tasks~\cite{wei2021finetuned}. However, the time complexity and costs of full-parameter fine-tuning can be exceedingly high due to the huge number of parameters in LLMs. To mitigate this issue, more efficient fine-tuning methods have been extensively developed, such as adapter tuning, prefix tuning, prompt tuning and LORA~\cite{ding2023parameter, li2021prefix, lester2021power, hu2021lora}.

\section{Methodology}
\label{sec:methodology}

\subsection{Problem Formulation}

\begin{figure}
\centering\includegraphics[width=\linewidth]{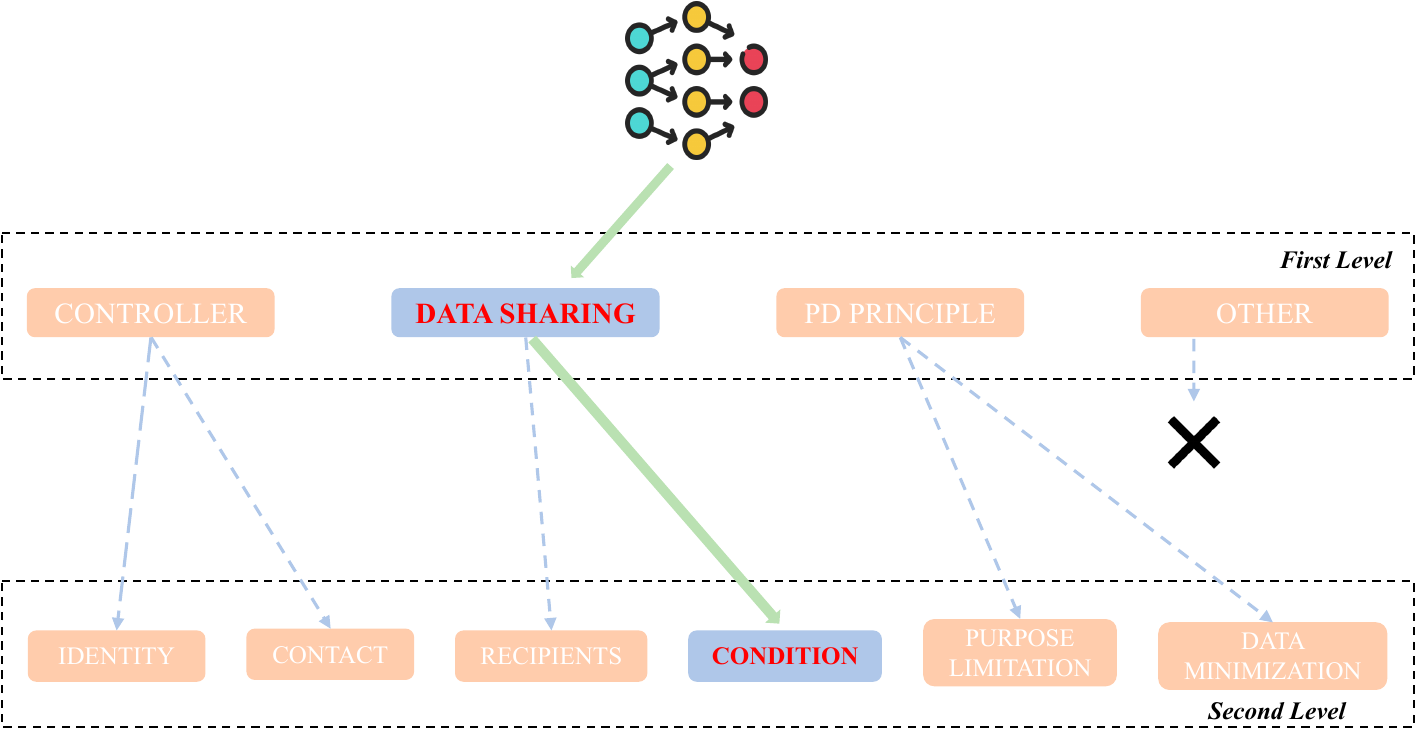}
\caption{A partial hierarchy of GoPPC-150.}
\label{fig:dag}
\end{figure}

The problem can be defined as a multi-class multi-label classification task of assigning a segment in a given privacy policy one or more concepts defined in a relevant taxonomy.
Among all privacy policy taxonomies, the one supporting the privacy policy corpus GoPPC-150~\cite{tang2024gdpr} is the most advance and the first multi-level one, with fine-grained privacy policy concepts especially those related to the GDPR. A partial hierarchy of the GoPPC-150 taxonomy is illustrated as a directed acyclic graph (DAG) in Figure~\ref{fig:dag}.

To elucidate the process of privacy policy concept classification, let us consider an illustrative example. Given a privacy policy segment $X$, the classifier $H(\cdot)$ produces a label $Y_1$ indicating one of the first-level nodes of the taxonomy, denoted as $Y_1 = H(X)$. A special value of $Y_1$ is `OTHER', indicating that $X$ does not match any first-level nodes in the taxonomy. If $Y_1$ does refer to a leaf node, such as `DATA SHARING' the partial taxonomy in Figure~\ref{fig:dag}, a subsequent classification task will proceed to determine the associated second-level node following a similar process, denoted as $Y_2 = H(X, Y_1)$, where $Y_2$ refers to the produced second-level node. The process can continue until a leaf node is reached, although for GoPPC-150 only the first two levels have sufficient data so the process will stop at the second level. The final classification result of $X$ is therefore a cascaded code denoted by $Y_1.Y_2$, e.g., `DATA SHARING.CONDITION'. Note that $X$ may be labeled multiple concepts so more than one final classification code could be produced.

\subsection{Design Prompts}
\label{subsec:methodology_prompts}

We explored applying prompt engineering to privacy policy analysis. Given that the OPP-115 privacy policy corpus represents the first published and most widely used dataset in this domain, we utilized it as the benchmark to assess the effectiveness of various prompt designs. To comprehensively assess the impact of different prompt engineering techniques, such as few-shot and CoT, we designed five different types of prompts to elicit related concepts from privacy policy segments.

Each type of prompts are designed to provide different levels of guidance and context to LLMs. Figure~\ref{fig:prompts} in Appendix~\ref{appendix_Prompts} shows greater details of the five types of prompts.

\begin{itemize}
\item Prompt 1 simply describes the task without providing any additional explanations or examples.

\item Prompt 2 extends Prompt 1 by including detailed explanations of all the 12 concepts defined in OPP-115.

\item Prompts 3 and 4, in addition to providing category explanations, are designed for few-shot learning with one (Prompt 3) or two (Prompt 4) examples for each category.

\item Prompt 5 introduces CoT as a structured reasoning process to guide LLMs through a step-by-step approach.
\end{itemize}


\subsection{Fine-Tuning}

\begin{figure} [!htb]
\centering
\includegraphics[width=\linewidth]{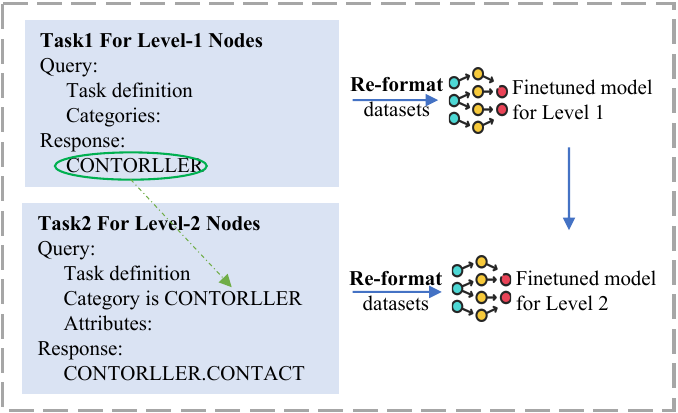}
\caption{Two-leveled fine-tuning process.}
\label{fig:finetune}
\end{figure}

We propose a method to streamline the adaptation of LLMs to hierarchical classification tasks, using the multi-level corpus GoPPC-150 as the benchmark dataset. The fine-tuning process involves two distinct tasks: predicting first-level nodes based on segment content, and subsequently predicting second-level nodes based on both segment content and the predicted first-level nodes. Figure~\ref{fig:finetune} shows the two-leveled fine-tuning process. The process is progressive, and LLMs acquire two-stage prediction capabilities through this process.

\section{Experiments and Results}
\label{sec:experiments_results}

\subsection{Setup}

\textbf{Corpora.} Different corpora employ different concept taxonomies. These taxonomies differ in both granularity, multi-level taxonomies being more detailed than single-level ones, and construction standard, with some tailored to specific regulations such as the GDPR and others being more general. OPP-115 uses a simple single-level taxonomy to broadly categorize concepts. The taxonomy it employs was defined based on real-world privacy policies by law experts over multiple iterations. In contrast, corpora like GoPPC-150 and APPCP-100 employ a multi-level taxonomy that is specifically designed to align with the hierarchical nature of privacy policies and a data protection regulatory framework, thereby offering greater relevance and applicability to real-world scenarios. For a comprehensive study of diverse taxonomies, we selected OPP-115 and GoPPC-150, complemented by their Chinese counterparts, CAPP-130 and APPCP-100, which adhere to a single-level and a multi-level taxonomy, respectively.

The OPP-115 corpus consists of 10 concept categories, with the final category `OTHER' further subdivided into three distinct categories: `Introductory/Generic', `Privacy Contact Information', and `Practice Not Covered'. We considered all the 12 categories as a single-level taxonomy for our experiments. The GoPPC-150 corpus has nodes at three levels, but only 14 first-level and 21 second-level nodes have sufficient data, which were used for our experiments. The CAPP-130 corpus classifies privacy policy segments by three aspects: importance, risk, and topic classification. This paper focuses on the topic classification task (with 11 topical categories) that is aligned with the classification of concepts. For APPCP-100, the nodes that appear infrequently are filtered and 13 first-level, 25 second-level, and 16 third-level nodes were selected for our experiments.

\textbf{Models.} To ensure that our experimental results align with the latest advancements of LLMs, we selected three widely recognized open-source models for our experiments. These include Llama developed by Meta~\cite{touvron2023llama}, the Qwen series developed by Alibaba~\cite{bai2023qwentechnicalreport}, and ChatGLM developed by the Tsinghua University~\cite{glm2024chatglm}. We also conducted limited experiments using GPT, the most studied closed-source LLM from Open AI~\cite{radford2018improving}. These models were chosen because of their strong performance in various benchmarks and relevance to current research trends. We aim to provide a robust and comprehensive evaluation of our proposed method using these models.

\subsection{Evaluation Metrics}

We treated our multi-label multi-class classification task as multiple independent binary classification tasks. Therefore, we employed classic metrics for binary classifiers such as precision, recall, and the F1 score. To comprehensively evaluate performance across all labels, we calculated the average of these metrics over all labels. The macro-average calculates the arithmetic mean of the metrics across all categories, treating each category equally regardless of its frequency. It provides the model's overall performance by evaluating its ability to discriminate across all categories independently. In contrast, the micro-average gives more weight to categories with a larger number of samples, effectively reflecting the model's performance relative to the actual distribution of categories. Both metrics are commonly employed to assess the model performance in multi-label classification tasks, and neither can be considered universally superior. Consequently, we compared and reported both macro-average and micro-average.

\subsection{Prompt-Based Experiments}
\label{subsec:prompt_experiments}

We conducted experiments to evaluate the effectiveness of prompt engineering using the Llama3-8B-Instruct model. Specifically, we tested five different types of prompts on the OPP-115 corpus, using a configuration with temperature of 0.6, top-p of 0.9, and top-k of 50. The experimental results are summarized in the Table~\ref{tab:prompt}, which includes F1 scores as well as macro- and micro-average scores.

\begin{table*}[!htb]
\centering
\small
\begin{tabular}{*{6}{c}}
\toprule
Label & Prompt 1 & Prompt 2 & Prompt 3 & Prompt 4 & Prompt 5\\
\midrule
First Party Collection/Use & 0.740 & 0.774 & 0.762 & 0.788 & 0.748\\
Third Party Collection/Use & 0.730 & 0.772 & 0.762 & 0.714 & 0.758\\
User Choice/Control & 0.366 & 0.441 & 0.465 & 0.458 & 0.478\\
User Access, Edit and Deletion & 0.533 & 0.582 & 0.667 & 0.646 & 0.611\\
Data Retention & 0.135 & 0.217 & 0.385 & 0.300 & 0.204\\
Data Security & 0.549 & 0.517 & 0.549 & 0.550 & 0.471\\
Policy Change & 0.472 & 0.467 & 0.512 & 0.568 & 0.532\\
Do Not Track & 0.240 & 0.300 & 0.286 & 0.222 & 0.240\\
International/Specific Audiences & 0.451 & 0.768 & 0.803 & 0.835 & 0.762\\
Introductory/Generic & 0.436 & 0.564 & 0.431 & 0.471 & 0.514\\
Privacy Contact Information & 0.731 & 0.696 & 0.707 & 0.682 & 0.714\\
Practice Not Covered & 0.091 & 0.198 & 0.250 & 0.220 & 0.193\\
\midrule
Macro Average & 0.456 & 0.525 & \textbf{0.548} & 0.538 & 0.519\\
Micro Average & 0.548 & 0.620 & \textbf{0.636} & 0.623 & 0.611\\
\bottomrule
\end{tabular}
\caption{Performance of Llama3-8B-Instruct using 5 types of prompts on the OPP-115 corpus (F1 scores).}
\label{tab:prompt}
\end{table*}

The overall performance of the model was relatively poor. The poor performance for some categories, like `Introductory/Generic' and `Practice Not Covered', can be attributed to them being less clearly defined, making it challenging to assess based on individual sentences.\footnote{As reported in \cite{opp-115}, during the annotation process, they have shown significant disagreement among the three legal experts. The `OTHER' class (which covers the two aforementioned concepts) has the poorest inter-rater agreement, with Fleiss' kappa equal to just 0.49.} Also, the model's strong hallucination led to an excessive number of false positives, resulting in high recall but low precision rates and consequently poor F1 scores, especially for categories like `Data Retention', `Data Security', and `Do Not Track'. For instance, a segment that mentions protecting user privacy, like ``We are committed to protecting and respecting your privacy'' was mistakenly classified as `Data Security'. While it refers to the commitment to privacy protection, it does not describe specific security measures, therefore, it is not related to this concept. On the other hand, the model performed well on categories such as `First Party Collection/Use', `Third Party Collection/Use', and `International/Specific Audiences', probably due to their being easier concepts.

Differences in performance were observed between the five types of prompts. Prompt 1, which only contains a task description, performed the worst, which is not surprising given it providing the least information. Prompt 2 adds explanations of concept categories, so the model can understand the concepts better. Prompts 3 and 4 show a significant improvement in a few-shot setting. Compared to Prompt 3, where one example per concept category is used, Prompt 4 includes two examples. However, the increase in the number of examples did not improve the results, which was unexpected, indicating more is not always better. Prompt 5 used the CoT approach, but it performed the second worst, which was also unexpected.

\subsection{Temperature Experiments}

Several factors, such as sampling methods, temperature, top-p and top-k, can significantly impact model performance. We conducted experiments to show the role of temperature. We employed Prompt 3 for these experiments because it achieved the best performance among all five types of prompts as reported in the previous subsection. Utilizing the Llama3-8B-Instruct model again, experiments were conducted on the OPP-115 corpus with top-p fixed at 0.9 and top-k set to 50, while the temperature was varied across 0.3, 0.6, and 0.9, including a greedy generation for comparison. The performance under different generation configurations are shown in Table~\ref{tab:temperature}. It indicates that the Llama3-8B-Instruct model exhibits limited sensitivity to temperature variations for the task of concern.

\begin{table}[!htb]
\centering
\small
\begin{tabular}{ccc}
\toprule
Setting & Macro Average & Micro Average\\
\midrule
Greedy & 0.536 & 0.629\\
T=0.3 & 0.546 & 0.632\\
T=0.6 & \textbf{0.548} & \textbf{0.636}\\
T=0.9 & 0.541 & 0.634\\
\bottomrule
\end{tabular}
\caption{Effect of temperature on the model performance (F1 scores).}
\label{tab:temperature}
\end{table}

\subsection{Fine-Tuning: Baseline Experiments}

We conducted experiments to evaluate the capability of fine-tuning the smaller versions of the three selected mainstream open-source LLMs, Llama3-8B, Qwen1.5-7B, and ChatGLM3-6B, utilizing four privacy policy corpora, OPP-115, GoPPC-150, CAPP-130, and APPCP-100. We perform LoRA fine-tuning on an RTX 4090 machine, primarily because LoRA significantly reduces computational costs compared to full fine-tuning and it can achieve a performance comparable to full fine-tuning in many scenarios, and it usually performs better than other alternative fine-tuning methods~\cite{hu-etal-2023-llm}.

For OPP-115, we selected the results of PrivBERT~\cite{Srinath_2021} as the baseline. For GoPPC-150, we adopted the PrivBERT+NN (neural network) approach used in~\cite{tang2024gdpr} as the baseline. For CAPP-130, we used RoBERTa as the baseline because it achieved best performance among all models as described in~\cite{wen2024capp}. For APPCP-100, we employed BERT+RF (random forests) described in~\cite{zhang2024appcp} as the baseline. 

Table~\ref{tab:comparsion} presents the performances of different LLMs compared with the baselines, demonstrating a significant improvement\footnote{We also conducted experiments using Llama3.1 and Qwen2.5 on OPP-115 corpus. But it shows no improvement compared to Llama3 and Qwen1.5. So we did not employ them on other corpora. Llama3.1-8B: macro F1 0.828, micro F1 0.871; Qwen2.5-7B: macro F1 0.825, micro F1 0.872.}. Notably, the advantages of LLMs are more pronounced on GoPPC-150, suggesting their superiority in handling such complex and fine-grained tasks. Llama3-8B demonstrates a superior performance on the two English corpora but a slightly lower performance on the two Chinese corpora, likely due to the limited coverage of Chinese in its pre-training corpus. Due to the importance of GPT series in the field of LLMs, we also experimented with it. We fine-tuned and evaluated gpt3.5-turbo-0125 on the OPP-115 corpus. The final performance, with a macro-average F1 score of 0.801 and a micro-average F1 score of 0.851, shows no improvement over other three open-source models. Due to the prohibitive costs of fine-tuning GPT and its lack of significant performance advantages compared to other LLMs, we strategically limited our experimental evaluation of GPT to a single corpus (OPP-115).

\begin{table*}[!htb]
\centering
\small
\begin{tabular}{*{14}{c}}
\toprule
\multicolumn{2}{c}{\multirow{2}{*}{Standard}} & \multicolumn{2}{c}{Baseline} & \hspace{0.5em} & \multicolumn{2}{c}{Llama3-8B} & \hspace{0.5em} & \multicolumn{2}{c}{Qwen1.5-7B} & \hspace{0.5em} & \multicolumn{2}{c}{Chatglm3-6B}\\
\cmidrule{3-4}\cmidrule{6-7}\cmidrule{9-10}\cmidrule{12-13}
& & macro & micro & & macro & micro & & macro & micro & & macro & micro\\
\midrule
OPP-115 & All & 0.830 & 0.870 & & \textbf{0.836} & \textbf{0.877} & & 0.831 & 0.868 & & 0.819 & 0.867\\
\midrule
\multirow{2}{*}{GoPPC-150} & Level 1 & 0.669 & 0.697 & & \textbf{0.717} & \textbf{0.725} & & 0.709 & 0.718 & & 0.705 & 0.712\\
& All & 0.529 & 0.589 & & \textbf{0.618} & \textbf{0.685} & & 0.612 & 0.673 & & 0.609 & 0.668\\
\midrule
CAPP-130 & Topic & 0.841 & 0.819 & & 0.838 & 0.821 & & 0.852 & 0.829 & & \textbf{0.858} & \textbf{0.837}\\
\midrule
\multirow{2}{*}{APPCP-100} & Level 1 & 0.832 & 0.867 & & 0.831 & 0.875 & & \textbf{0.843} & \textbf{0.883} & & 0.840 & 0.878\\
& All & 0.767 & 0.846 & & 0.767 & 0.858 & & \textbf{0.782} & \textbf{0.865} & & 0.778 & 0.858\\
\bottomrule
\end{tabular}
\caption{Performance of different LLMs with fine-tuning compared with baseline.}
\label{tab:comparsion}
\end{table*}

\subsection{Fine-Tuning: Experiments on Different Model Sizes}

Prior research~\cite{wei2022abilities} has indicated that, for some tasks, larger LLMs exhibit a significantly superior performance compared to smaller ones. However, in certain tasks, smaller models have been observed to achieve a substantial portion of the performance of larger models, thereby offering a more cost-effective and practical alternative. We conducted experiments to investigate this phenomenon for the task we are studying. We focused on the Qwen1.5 series, which encompasses a more diverse range of model scales especially at the lower end, including 0.5B, 1.8B, 4B, and 7B parameters. Qwen1.5-7B has also demonstrated a robust performance in our own experiments and other researchers' past studies, making it a reasonable choice.

\begin{figure}[!htb]
\includegraphics[width=0.48\linewidth]{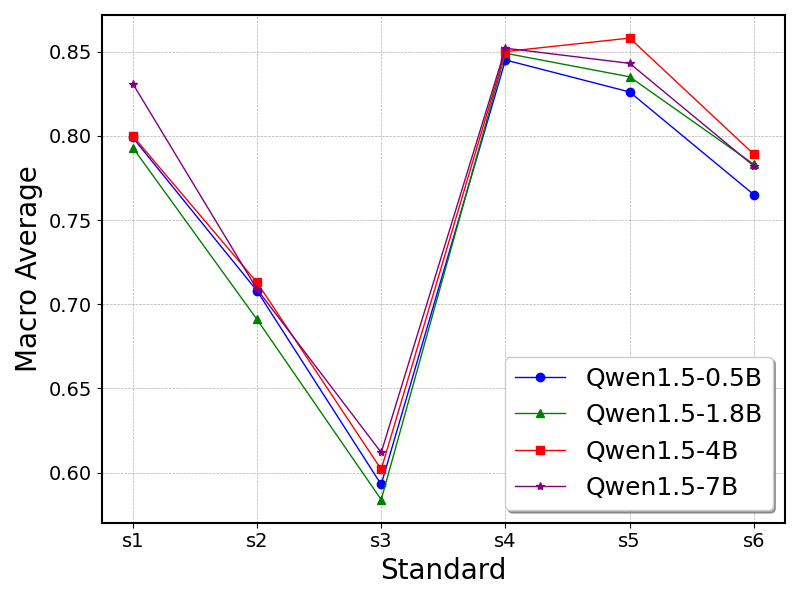} \hfill
\includegraphics[width=0.48\linewidth]{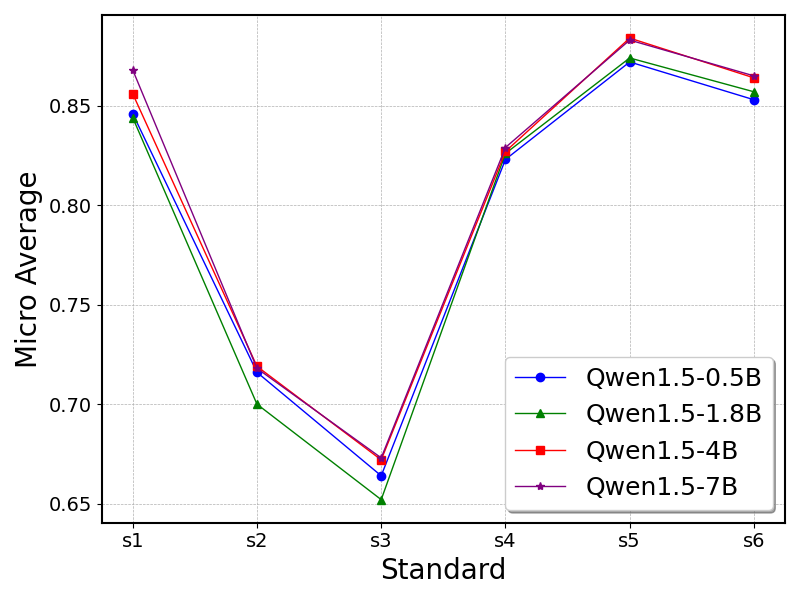}
\caption {Effect of the model size on the performance. Standards s1-s6 represent OPP-115, GoPPC-150 level-1 nodes, GoPPC-150 all nodes, CAPP-130, APPCP-100 level-1 node, and APPCP-100 all nodes, respectively.}
\label{fig:size}
\end{figure}

We evaluated the performance of Qwen1.5 models of various scales (0.5B, 1.8B, 4B, and 7B) on all the four privacy policy corpora. Figure~\ref{fig:size} shows the performances of Qwen1.5 models with different size, revealing a trend that larger models generally outperform smaller ones, but the improvement is small or marginal. Notably, after fine-tuning, the 0.5B model was already able to achieve over 90\% of the performance of the 7B model. Therefore, if the performance requirements are not high, smaller models have certain advantages.

\subsection{Fine-Tuning: Experiments on Single- vs Multi-Task Settings}

Single-task training focuses on optimizing a model for one specific task, while multi-task training involves training a model on multiple tasks simultaneously. We conducted experiments to explore the application of multi-task fine-tuning to enhance the ease of model deployment. Specifically, we integrated the first- and second-level node classification tasks of GoPPC-150 into a unified task framework by merging the two aforementioned single-task fine-tuning corpora. By fine-tuning the Llama3-8B model on this consolidated corpus, we achieved an excellent performance on both tasks simultaneously. Table~\ref{tab:multi-task} presents the performances of the two training paradigms. The performance of multi-task fine-tuning shows just a small significant drop compared to single-task fine-tuning, demonstrating its feasibility.

\begin{table}[!htb]
\centering
\small
\begin{tabular}{ccc}
\toprule
Method & Macro average & Micro average\\
\midrule
Single-task & 0.618 & 0.685\\
Multi-task & 0.614 & 0.679\\
\bottomrule
\end{tabular}
\caption{Performances of single- and multi-task paradigms.}
\label{tab:multi-task}
\end{table}

\subsection{Performance Comparison}

As mentioned in Section~\ref{sec:related_work}, \citet{goknil2024privacy} explored the use of prompt engineering and LLMs for automated privacy policy analysis, using OPP-115 as the corpus. To compare with their work, we adopted their best results on Llama3-8B. Since they did not consider the last three categories in OPP-115, we also excluded them in our performance comparison experiments, leading to slight discrepancies with the results reported in Section~\ref{subsec:prompt_experiments}. Table~\ref{tab:comparison_performance} presents a comparison of the performance figures (F1 scores) on the OPP-115 corpus. The results show that prompt engineering methods generally perform more poorly, while the fine-tuning method we employed demonstrates a far superior performance.

\begin{table}[!htb]
\centering
\small
\begin{tabular}{p{2.5cm}<{\centering}p{1cm}<{\centering}p{1cm}<{\centering}p{1.5cm}<{\centering}}
\toprule
Category & Goknil et al.'s & Ours (PE) & Ours (Finetuned)\\
\midrule
First Party Collection/Use & 0.760 & 0.789 & \textbf{0.939}\\
Third Party Sharing/Collection & 0.710 & 0.789 & \textbf{0.935}\\
User Choice/Control & 0.630 & 0.477 & \textbf{0.847}\\
User Access, Edit and Deletion & 0.730 & 0.667 & \textbf{0.821}\\
Data Retention & 0.400 & 0.348 & \textbf{0.696}\\
Data Security & 0.740 & 0.568 & \textbf{0.873}\\
Policy Change & 0.880 & 0.585 & \textbf{0.973}\\
Do Not Track & 0.810 & 0.300 & \textbf{1.000}\\
International and Specific Audiences& 0.810 & 0.827 & \textbf{0.918}\\
Micro Average & 0.730 & 0.694 & \textbf{0.916}\\
\bottomrule
\end{tabular}
\caption{Comparison of performances (F1 scores) on the OPP-115 corpus (PE = prompt-engineered).}
\label{tab:comparison_performance}
\end{table}

\section{Explainability}
\label{sec:explainability}

Compared to traditional deep learning methods, LLMs has the potential to offer enhanced explainability due to its capability of producing human-like texts in natural languages. To investigate the explainability of LLMs for privacy policy concept classification, we fed privacy policy segments along with their corresponding concept categories into an LLM, prompting it to analyze and explain the classification results. Specifically, our prompts include the task description, the concept categories' descriptions, the required output format, and some examples. The task description and concept categories' descriptions are consistent with those detailed in Section~\ref{subsec:methodology_prompts}. We instructed the LLM to, for each category, first explain its meaning and then analyze the segment's relevance to the category.

As an example, we utilized the Llama3-8B-Instruct model, with settings of temperature=0.6 and top-p=0.9, to generate explanations for 100 privacy policy segments randomly selected from OPP-115. We focused on the first 11 categories of the OPP-115 taxonomy, excluding the `Practice Not Covered' category, which does not require any specific explanation.

We established three metrics to assess the quality of the LLM-generated explanations, as explained below. Each metric was scored by three human annotators on a scale of 1, 2, or 3, where 1 indicates `poor performance', 2 indicates `acceptable performance', and 3 indicates `outstanding performance'.

\textbf{Completeness} assesses whether the explanation covers all key points of the privacy policy segment that identifies the relevant concept categories.

\textbf{Logicality} evaluates the accuracy of the model's understanding of the privacy policy segment and the coherence of the model's reasoning.

\textbf{Comprehensibility} focuses on the clarity and understanding of the explanation itself, especially in terms of language.

The three human annotators are three co-authors of the paper, all postgraduate research students, who conducted a qualitative evaluation of the explanations of LLM outputs based on the three metrics mentioned above. To prevent potential positive scoring bias, we included 10 decoy explanations that were made blind to the annotators. These decoy explanations were crafted to exhibit at least one aspect of relatively poor performance while maintaining basic explanatory quality. After aggregating the scores, the average scores for the three metrics are presented in Table~\ref{tab:explainscore}. The average scores of the 10 artificially crafted explanations are significantly lower across all three metrics compared to the LLM-generated ones. We assessed the inter-rater reliability among three annotators using Fleiss' kappa~\cite{Landis1977TheMO}. The results indicated a substantial agreement for all three metrics: 0.765 for completeness, 0.695 for logicality, and 0.656 for comprehensibility. Our primary finding is that LLMs exhibit very very good explainability in explaining the classification results of the 100 privacy policy segments, across all three metrics. Notably, the Llama3-8B-Instruct model tend to offer comprehensive analyses of the original text, which contributes to their strong performance in terms of completeness. Moreover, the language style of the LLM-generated content can be easily set to be clear, concise, and easy to understand through the use of prompts, thus demonstrating strong comprehensibility. However, the Llama3-8B-Instruct model' understanding of privacy policy segments occasionally lacks depth, which results in slightly lower logicality score.

\begin{table}[!htb]
\centering
\small
\begin{tabular}{cccc}
\toprule
Source & C1 & L & C2\\
\midrule
LLM & 2.84 & 2.73 & 2.87\\
Artificial & 2.43 & 2.10 & 2.73\\
\bottomrule
\end{tabular}
\caption{Average scores of LLM-generated explanations and artificially crafted explanations (C1 = completeness, L = Logicality, C2 = comprehensibility).}
\label{tab:explainscore}
\end{table}

\section{Conclusion}

This paper proposes a method for utilizing LLMs to classify concepts in a privacy policy based on an established taxonomy. Unlike prior studies, we provided a comprehensive evaluation of LLMs in this domain, incorporating both prompt engineering and LoRA techniques, and assess performance across four SOTA privacy policy corpora and multiple mainstream LLMs, achieving SOTA results against existing methods. We investigated the effects of factors such as temperature, model size, and training paradigm. To enhance the explainability of the classification results, we used LLMs to generate explanations for the identified concepts and designed an evaluation framework to assess the explanations based on three metrics: completeness, logicality, and comprehensibility. The findings demonstrate that LLMs can provide satisfactory explanations to three human annotators. This paper highlights the great potential of LLMs for both automating analysis of privacy policies and producing useful human-understandable explanations, therefore opening up their use for many downstream tasks in this important application domain.

\section{Limitations}

The performance achieved using prompt engineering in our experiments is quite poor. This can be two reasons: more advanced prompt engineering methods are necessary, and LLMs may not have seen enough privacy policies so fine-tuning is a must to improve the performance of any tasks about privacy policies. We call for more follow-up research to clarify both points.

Due to resource limitation, we primarily utilized smaller and locally deployed models. While these models achieved SOTA performances in our experiments and demonstrated a great potential, we did not experiment with larger models like Llama-3-70B and GPT-4. As a result, we were unable to evaluate the performance upper bound of LLMs for privacy policy concept classification. Additionally, we achieved promising results using LoRA. Prior studies~\cite{hu2021lora} have shown that LoRA can closely approximate the performance of full-parameter fine-tuning. However, the underlying mechanisms of these approaches differ, and further research is required to fully explore the potential of full-parameter fine-tuning.

LLMs also benefit from pre-training to acquire domain-specific knowledge. Some studies~\cite{gupta2023continual, ke2023continual} adopted the continual pre-training paradigm, enabling models to perform unsupervised learning on domain-specific corpora before being fine-tuned for specific tasks. This approach allows LLMs to acquire substantial knowledge in a given domain and therefore likely to be able to solve targeted problems more effectively. 
In this paper, we did not adopt the continual pre-training paradigm, but relied on fine-tuning to help LLMs learn domain-specific knowledge. The effectiveness of the continual pre-training paradigm remains an area for future research.

In order to support other researchers to reproduce our results and to conduct follow-up research, we will make all data and code used publicly available.

\bibliography{main}

\appendix

\section{Details of Five Types of Prompts}
\label{appendix_Prompts}

\begin{figure*}[!htb]
\includegraphics[width=\linewidth]{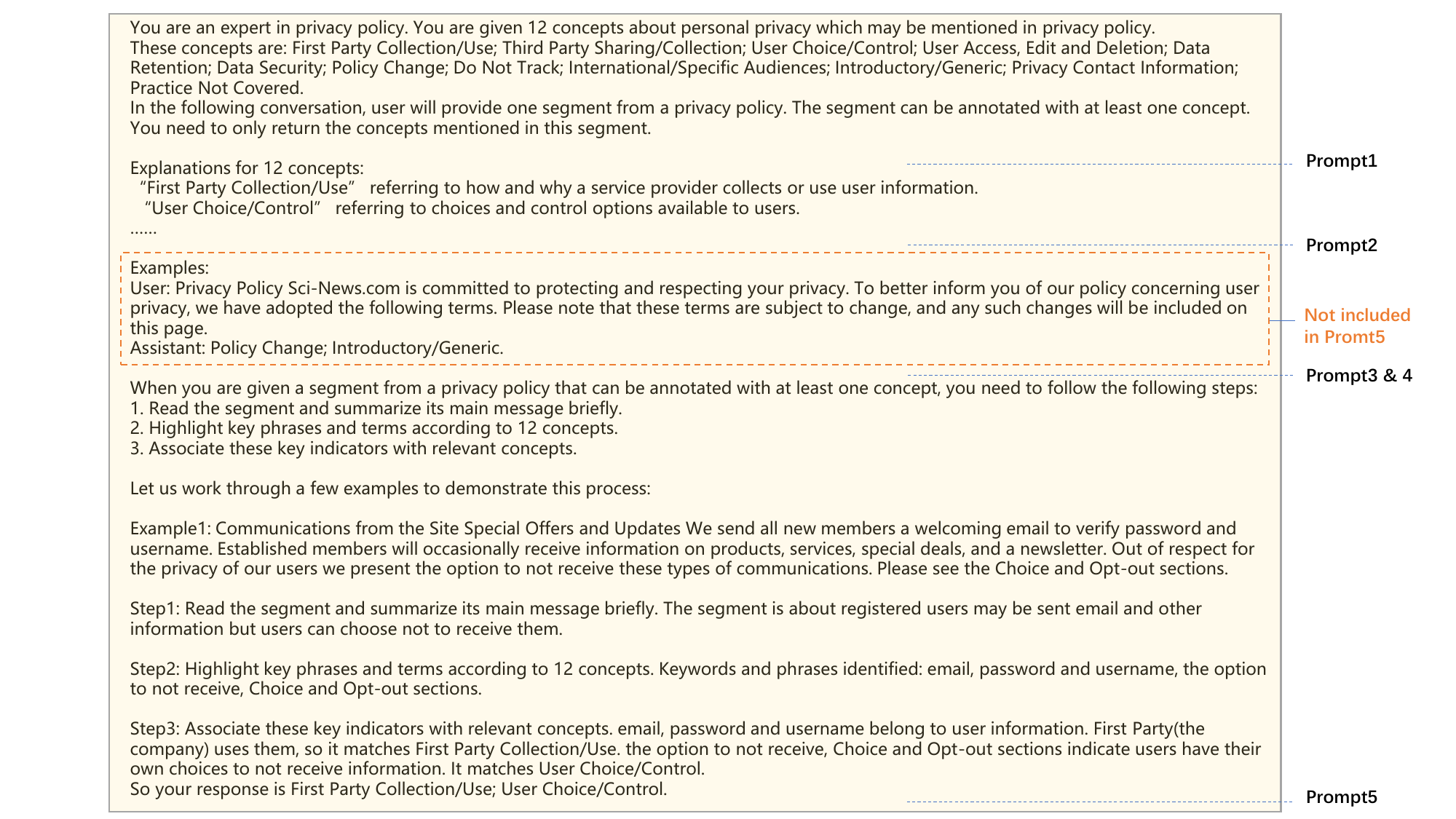}
\caption{Examples of the five types of prompts used in our experiments. The content within the yellow dashed box is not included in Prompt 5.}
\label{fig:prompts}
\end{figure*}

As mentioned in Section~\ref{subsec:methodology_prompts}, we designed five types of prompts, and Figure~\ref{fig:prompts} shows an example of each of the five types of prompts.

\end{document}